\documentclass{article}
\usepackage{spconf,amsmath,graphicx}
\usepackage{amssymb}
\usepackage{booktabs}
\usepackage{url}
\usepackage{enumitem}

\makeatletter
\def\bstctlcite{\@ifnextchar[{\@bstctlcite}{\@bstctlcite[@auxout]}}
\def\@bstctlcite[#1]#2{\@bsphack
  \@for\@citeb:=#2\do{%
    \edef\@citeb{\expandafter\@firstofone\@citeb}%
    \if@filesw\immediate\write\csname #1\endcsname{\string\citation{\@citeb}}\fi}%
  \@esphack}
\makeatother
\title{Evaluating the Diagnostic Contribution of Learned Myocardial Motion Beyond Volumetric Indices in Cine MRI}
\name{Chou Mo$^{1}$, Shiyi Wang$^{2}$, Ruochen Sun$^{2}$, Fangxu Xing$^{2}$}
\address{$^{1}$Department of Mathematics, University of California Los Angeles, Los Angeles, USA \\
         $^{2}$Department of Radiology, Harvard Medical School, Boston, USA}

\begin{document}
\bstctlcite{IEEEexample:BSTcontrol}
\ninept
\maketitle

\begin{abstract}
Recent deep learning models for cardiac diagnosis from cine MRI encode myocardial motion and measure its benefit by classification accuracy on the ACDC benchmark. However, ACDC defines its diagnostic classes by thresholds on ejection fraction (EF), ventricular volume, and wall thickness, and among the disease classes, only infarction also requires abnormal myocardial contraction. Therefore, a motion representation can improve ACDC accuracy only by adding label information that these indices do not capture. We test this with DyMRL (which combines diffeomorphic registration, a learned motion encoder, a selective state-space model with optional cardiac-phase conditioning, and eight clinical indices) against a clinical-only model on the same folds and seeds, with last-epoch evaluation and paired tests. Clinical indices alone reach $92.5\%$ four-class accuracy. Across two motion encoders and two phase-conditioning mechanisms, no configuration with learned motion exceeds this (joint fusion $82.5$--$85.0\%$, late fusion $87.9$--$91.7\%$, motion alone $50.0$--$61.2\%$). For infarction versus dilated cardiomyopathy, pre-specified regional strain features add $+2.5$ points [$0.0$, $+7.5$], which is not statistically significant ($p=1.00$), and correlate with EF (Spearman $|\rho|$ up to $0.78$). Selecting the epoch by held-out-fold accuracy picks an epoch before the registration network is trained in 11--13 of 15 folds (3-class, with clinical indices), so high accuracy need not come from learned motion. Therefore, accuracy on volumetrically defined labels cannot separate the contribution of motion from that of the indices.
\end{abstract}

\begin{keywords}
Cine MRI, cardiac motion, myocardial strain, multimodal fusion, evaluation protocol, domain shift
\end{keywords}

\section{Introduction}
\label{sec:intro}
Myocardial deformation carries diagnostic information \cite{bijnens2009velocity}. Strain abnormalities can appear before EF declines \cite{mihos2025speckle}, and feature-tracking global longitudinal strain predicts mortality in patients with preserved EF, independently of clinical and imaging risk factors \cite{romano2020feature}. In both cases, deformation carries information that EF does not, in patients whose EF is normal or not yet abnormal. Across the EF range, global longitudinal and circumferential strain correlate with CMR EF at $r=-0.85$ and $r=-0.95$ \cite{onishi2015global}.

On the Automated Cardiac Diagnosis Challenge (ACDC) benchmark \cite{bernard2018deep}, each disease class is defined by thresholds on volumetric and functional indices, and subjects with ambiguous indices were excluded:
\begin{itemize}[leftmargin=*, nosep]
\item Dilated cardiomyopathy (DCM): EF $<40\%$, LV volume $>100$\,mL/m$^2$, and diastolic wall thickness $<12$\,mm;
\item Hypertrophic cardiomyopathy (HCM): EF $>55\%$ and myocardial segments thicker than $15$\,mm in diastole;
\item Abnormal right ventricle (RV): RV volume $>110$\,mL/m$^2$ (men) or $>100$\,mL/m$^2$ (women), and/or RV EF $<40\%$; and
\item Previous myocardial infarction (MINF): EF $<40\%$ and abnormal myocardial contraction.
\end{itemize}
Consistent with these definitions, classifiers on volume, thickness, and shape features from segmentations reach $92\%$ \cite{isensee2017automatic} to $100\%$ \cite{khened2019fully} accuracy on the 50-subject test set. Because EF summarizes global contraction, such features already contain the global component of motion. A motion representation can therefore raise accuracy above a model given these features only by supplying (a)~regional or temporal information that a class definition depends on, which among the ACDC disease classes is the case for MINF alone, or (b)~an index that the features omit. Models that encode motion through apparent flow \cite{zheng2019explainable} or deformation fields \cite{liu2023successive} are evaluated on these labels, and their accuracy measures the contribution of motion only relative to a model given the same features on the same folds. In 1{,}337 patients, adding longitudinal strain features to morphological and functional features improved most classification metrics by only $0.001$--$0.03$ \cite{jacob2025deep}. Outside ACDC, motion representations have improved survival prediction \cite{bello2019deep}, an outcome not defined by volumetric indices.

\section{Method}
\label{sec:method}
\begin{figure*}[t]
    \centering
    \includegraphics[width=0.85\textwidth]{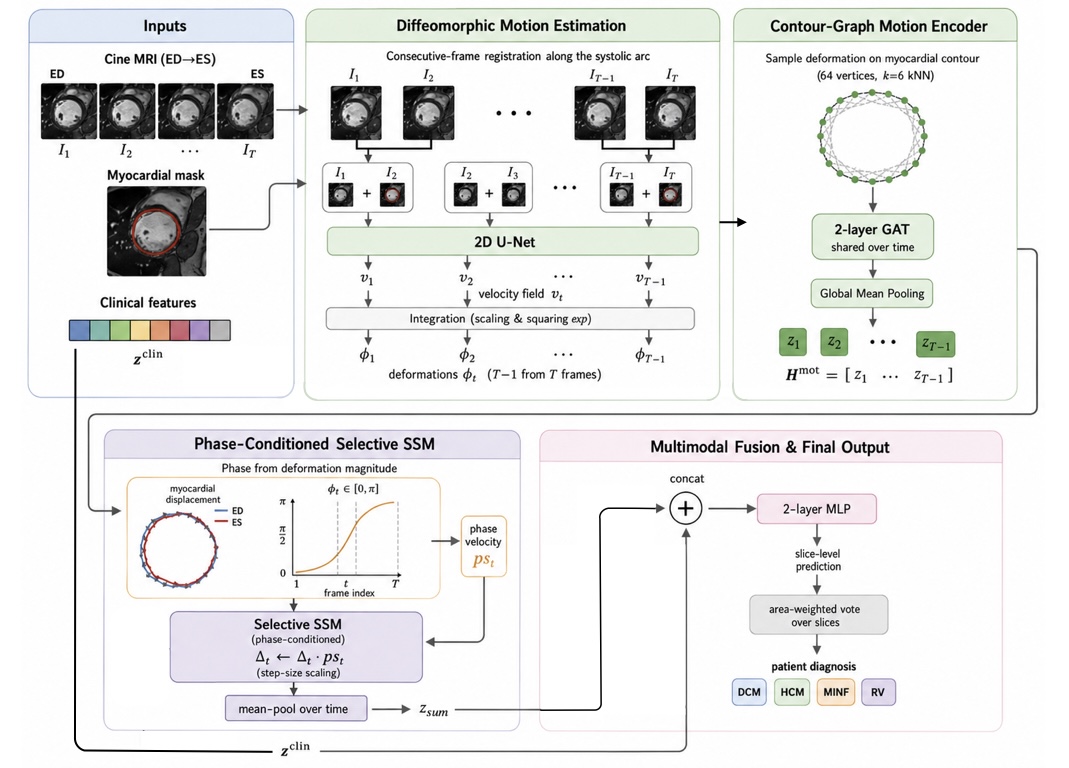}
    \caption{Overview of DyMRL. Consecutive systolic frame pairs are registered by a diffeomorphic U-Net (Stage~1) into a deformation field $\varphi$. A GAT on the myocardial contour graph summarizes motion into a per-pair embedding $z_t$ (Stage~2), which a phase-conditioned selective SSM aggregates into $z_{\mathrm{sum}}$ (Stage~3), with the phase derived from $\varphi$. Stage~4 fuses $z_{\mathrm{sum}}$ with eight clinical indices through an MLP. Per-slice logits are combined by myocardium-area-weighted voting into the patient-level diagnosis.}
    \label{fig:pipeline}
\end{figure*}
DyMRL comprises four jointly trained stages (Fig.~\ref{fig:pipeline}). Each input is a 2D short-axis slice sampled at every second frame from end-diastole (ED) to end-systole (ES) and resized to $128\times128$. ES is included only when ES$-$ED is even (48 of 100 ACDC subjects), and otherwise the sequence ends one frame before ES. The ED myocardium mask is used for all frames.

\subsection{Stages 1--3: Motion Estimation, Encoding and Temporal Model}
\label{sec:reg}
For consecutive frames $I_f$ (fixed) and $I_m$ (moving), a 2D U-Net predicts a stationary velocity field $v$, integrated into $\varphi=\exp(v)$ by seven scaling-and-squaring steps \cite{dalca2018unsupervised} (the output head is zero-initialized, thus $\varphi$ starts at the identity). Registration is trained with a local normalized cross-correlation $L_{\mathrm{ncc}}$ ($9\times9$ windows, restricted to the myocardium), a bending energy $L_{\mathrm{bend}}$ on $v$, and a folding penalty $L_{\mathrm{fold}} = \mathrm{ReLU}(-\det J_\varphi)$.

\textbf{Contour-graph encoder.} The largest external contour of the ED myocardium mask (the epicardial boundary) is resampled by arc length to $N=64$ vertices, and the same vertices are used for every frame pair $t$. Each vertex carries its displacement, sampled from $\varphi_t$, and the geometry of its ED position (coordinates, centroid distance, and polar angle). A $k$-nearest-neighbor graph ($k=6$) is encoded by a two-layer graph attention network \cite{velickovic2017graph}, and mean pooling over vertices gives $z_t\in\mathbb{R}^{64}$.

\textbf{CNN encoder (ablation).} Without a contour, a four-layer CNN reads $[\varphi_x, \varphi_y, \text{ED mask}]$ over the whole image, followed by global mean pooling.

\textbf{Selective SSM.} The frame-pair embeddings $z_1,\dots,z_T$ are processed by a selective state-space model \cite{gu2023mamba} with input-dependent step size $\Delta_t=\mathrm{softplus}(\mathrm{Linear}(z_t)+b)$, and the outputs are mean-pooled into $z_{\mathrm{sum}}\in\mathbb{R}^{64}$.

\textbf{Phase conditioning.} Let $d_t$ be the mean displacement magnitude of $\varphi_t$ over the ED mask. We define a phase velocity $\mathrm{ps}_t$ and a phase position $\psi_t$:
\begin{equation}
\mathrm{ps}_t = \tilde T\,d_t/\sum_k d_k, \qquad
\psi_t = \pi\,\frac{\sum_{k\le t} d_k - \tfrac12 d_t}{\sum_k d_k},
\end{equation}
where $\tilde T$ is the padded sequence length and $\mathrm{ps}_t$ is clipped to $[0.1, 5]$. \texttt{dt\_scale} replaces $\Delta_t$ by $\Delta_t\,\mathrm{ps}_t$; \texttt{dt\_sincos} adds a zero-initialized projection of $[\sin\psi_t, \cos\psi_t]$ to the step bias $b$. As controls, \texttt{dt\_scale} is also run with $\mathrm{ps}_t$ permuted in time (\emph{shuffled}) or replaced by a per-sequence constant (\emph{constant}).

\subsection{Stage 4: Fusion and Training}
\label{sec:multimodal}
Eight clinical indices are computed from the ED and ES masks and metadata: EF, LV end-diastolic and end-systolic volume (EDV, ESV), myocardial volume, RV volume, RV/LV volume ratio, height and weight, scaled by fixed constants. They are concatenated with $z_{\mathrm{sum}}$ and mapped to class logits by an MLP (72$\to$256$\to C$, dropout 0.5). The loss is $L = L_{\text{cls}} + 0.1\, L_{\text{ncc}} + 0.01\,(L_{\text{bend}} + L_{\text{fold}})$, with $L_{\text{cls}}$ a focal loss ($\gamma = 2$) \cite{8417976} with label smoothing 0.1. The registration network is frozen for the first 20 epochs ($\varphi$ is the identity during these epochs). Patient logits are the mean of slice logits weighted by ED myocardium area.

\subsection{Comparison Models}
\label{sec:baselines}
Four comparison models isolate the contribution of motion, where each removes one alternative explanation for a result.
(i)~\textbf{Clinical-only} measures what the eight indices achieve alone. We adopt the Stage~4 MLP with the motion input $z_{\mathrm{sum}}$ removed, trained with the same loss and optimizer, with each patient's index vector repeated once per slice so that it is seen as often as in DyMRL. We also fit a logistic regression on standardized indices.
(ii)~\textbf{Motion-only} measures the label information in the learned motion representation. We adopt DyMRL with the clinical indices set to zero.
(iii)~\textbf{Late fusion} tests whether joint training prevents motion from contributing. We collect, per seed, the mean of the softmax outputs of the separately trained motion-only and clinical-only MLP models.
(iv)~\textbf{Regional strain features} test whether global pooling or the learned encoder discards regional information. They require no training and were specified before any label was used. Each frame from ED to ES is registered to ED, ensuring displacements are Lagrangian (relative to ED). Registration is per-slice optimization (1.5\,mm in-plane, $32\times32$ stationary velocity grid, local NCC within the dilated ED LV and myocardium plus $0.1\times$ bending energy, coarse to fine). Green--Lagrange strain is projected onto radial and circumferential directions and averaged in six ED myocardial sectors anchored at the RV centroid. From the sector strains, we compute six patient features, each a myocardium-weighted mean over slices: global circumferential and radial strain (GCS, GRS; sector means of peak strain), the sector s.d.\ of each, the worst-sector circumferential deficit, and the sector s.d.\ of time to peak circumferential strain. On a synthetic phantom with known strain, the recovered sector circumferential strain was within $0.01$ of the true value. The features enter a logistic regression alone and appended to the clinical indices.

\section{Experiments and Results}
\label{sec:experiments}

\subsection{Experimental Setup}
\label{sec:setup}
\textbf{Data.} We use the 100 labeled ACDC training subjects \cite{bernard2018deep}, 20 per class: normal (NOR), DCM, HCM, MINF and RV. We evaluate four tasks: 5-class (all 100 subjects), 4-class disease (DCM, HCM, MINF, RV; 80), 3-class (DCM, HCM, NOR; 60) and DCM vs MINF (40). The 3-class task contains the classes shared with M\&Ms \cite{campello2021multi} and provides the models for zero-shot transfer. From M\&Ms, we use the 141 labeled subjects in these classes (DCM 54, HCM 49, NOR 38), where 67 are scanned on Siemens (the ACDC vendor) and 74 on Philips.

\textbf{Protocol.} 5-fold stratified, patient-level cross-validation with seeds 42, 7 and 2024 (clinical-only and strain models also 1 and 2), and all models share folds per seed. We report the last epoch. No separate validation split is used. Motion models train for up to 120 epochs with early stopping on the held-out-fold loss (patience 30), which uses the test fold and ended training early in 369 of 390 folds. The clinical MLP trains for all 120 epochs and the logistic regressions use no stopping rule, so this leakage affects only the motion models and is expected to favor them. Paired comparisons use seed-averaged class probabilities, with 95\% bootstrap CIs and two-sided permutation tests (5{,}000 each). For M\&Ms, the five fold models of each ACDC 3-class run are ensembled (mean logits) without fine-tuning, and the metric is balanced accuracy. Height is missing for 81 of 141 M\&Ms subjects (set to zero), so the M\&Ms clinical baseline omits height and weight. Training: AdamW (learning rate $10^{-4}$, weight decay $5\times10^{-4}$), cosine annealing, gradient clipping 1.0, batch size 8.

\begin{table}[t]
\centering
\caption{ACDC accuracy (\%), 5-fold CV, last epoch, mean\,$\pm$\,s.d.\ over seeds 42/7/2024 ($^\dagger$5 seeds). Strain rows use logistic regression and are compared with clinical-only logistic regression. On the 3 seeds shared with the motion models, the clinical-only MLP reaches $95.6\pm1.0$ / $92.9\pm0.7$ / $87.7\pm1.5$, and late fusion and all paired tests use these 3 seeds. \texttt{dt\_sincos} rows are given in the text.}
\label{tab:acdc}
\setlength{\tabcolsep}{3pt}
\resizebox{\columnwidth}{!}{%
\begin{tabular}{llccc}
\toprule
Inputs & Model & 3-class & 4-class & 5-class \\
\midrule
Clinical & MLP$^\dagger$                 & 96.0\,$\pm$\,0.9 & 92.5\,$\pm$\,0.9 & 88.4\,$\pm$\,1.5 \\
         & Logistic regression$^\dagger$ & 96.3\,$\pm$\,0.7 & 90.0\,$\pm$\,1.2 & 88.2\,$\pm$\,0.8 \\
\midrule
Motion   & Graph, no phase            & 76.1\,$\pm$\,5.9 & 55.4\,$\pm$\,4.0 & 42.3\,$\pm$\,10.0 \\
         & Graph, \texttt{dt\_scale}  & 88.9\,$\pm$\,2.5 & 50.0\,$\pm$\,4.5 & 38.3\,$\pm$\,10.2 \\
         & CNN, no phase              & --               & 61.2\,$\pm$\,5.7 & -- \\
         & Regional strain$^\dagger$  & --               & 63.0\,$\pm$\,2.4 & 54.0\,$\pm$\,2.2 \\
\midrule
Motion + & Joint, graph, no phase            & 91.7\,$\pm$\,1.7 & 82.5\,$\pm$\,4.5 & 81.7\,$\pm$\,2.1 \\
clinical & Joint, graph, \texttt{dt\_scale}  & 90.0\,$\pm$\,4.4 & 85.0\,$\pm$\,2.5 & 75.0\,$\pm$\,5.6 \\
         & Joint, CNN, no phase              & --               & 84.2\,$\pm$\,3.1 & -- \\
         & Late, graph, no phase             & 93.9\,$\pm$\,1.9 & 90.8\,$\pm$\,1.9 & 84.3\,$\pm$\,1.5 \\
         & Late, graph, \texttt{dt\_scale}   & 95.0\,$\pm$\,1.7 & 91.7\,$\pm$\,1.4 & 86.0\,$\pm$\,2.6 \\
         & Late, CNN, no phase               & --               & 88.3\,$\pm$\,1.9 & -- \\
         & Strain + clinical$^\dagger$        & --               & 90.5\,$\pm$\,1.9 & 84.8\,$\pm$\,0.8 \\
\bottomrule
\end{tabular}}
\end{table}

\subsection{ACDC: Impact of Learned Motion Beyond Clinical Indices}
\label{sec:acdc}
No configuration with learned motion exceeds the clinical-only MLP in mean accuracy (Table~\ref{tab:acdc}; \texttt{dt\_sincos}, 3/4/5-class: motion-only $73.9$/$53.8$/$50.3$, joint $89.4$/$84.6$/$77.7$, late fusion $94.4$/$87.9$/$83.3\%$). On paired, seed-averaged 4-class predictions, the joint models differ from clinical-only by $-8.8$ to $0.0$ points, late fusion by $-1.2$ to $+1.2$, and motion-only by $-18.8$ to $-26.2$ ($p \le 0.001$). In no task does a joint or late-fusion CI lie above zero. Regional strain added to clinical logistic regression changes accuracy by $+0.5$ (4-class) and $-3.4$ points (5-class). The CNN encoder changes motion-only accuracy by $+5.0$ points [$-6.2$, $+16.2$] ($p=0.53$) relative to the graph. Phase conditioning has no consistent effect: \texttt{dt\_scale} minus no phase, with clinical indices, is $+3.7$ [$-2.5$, $+11.3$] (4-class) and $-2.0$ [$-8.0$, $+4.0$] (5-class), and its 3-class motion-only gain does not recur in the 4- and 5-class tasks (Table~\ref{tab:acdc}).

\textbf{Epoch selection.} Selecting, for each fold, the epoch with the highest held-out-fold accuracy raises accuracy by $1.7$--$22.5$ points per run relative to the last epoch. With clinical indices on the 3-class task, the selected epoch is one of the first 20 in 11--13 of the 15 folds (5 folds $\times$ 3 seeds) of each configuration. During these epochs the registration network is frozen and $\varphi$ is the identity, so the motion encoder receives zero displacement. In those folds, the reported accuracy comes from the clinical indices and the ED contour geometry, so high accuracy under this rule does not by itself show that a model has learned motion.

\begin{table}[t]
\centering
\caption{Targeted test: DCM vs MINF (ACDC, $n=40$). Accuracy (\%) mean\,$\pm$\,s.d.\ over seeds 42/7/2024; seed-averaged accuracy [95\% CI]; paired difference from clinical-only with the same classifier type [95\% CI].}
\label{tab:dm}
\setlength{\tabcolsep}{2.5pt}
\resizebox{\columnwidth}{!}{%
\begin{tabular}{lccc}
\toprule
Model & Acc. & Seed-avg.\ acc. & $\Delta$ vs clinical \\
\midrule
Clinical-only, MLP                  & 89.2\,$\pm$\,2.9 & 87.5 [77.5, 97.5] & -- \\
Clinical-only, logistic             & 87.5\,$\pm$\,2.5 & 87.5 [77.5, 97.5] & -- \\
Learned motion-only (graph)         & 59.2\,$\pm$\,8.0 & 60.0 [45.0, 75.0] & -- \\
Learned joint (graph, MLP)          & 80.8\,$\pm$\,1.4 & 85.0 [72.5, 95.0] & $-2.5$ [$-12.5$, $+7.5$] \\
Regional strain only, logistic      & 89.2\,$\pm$\,3.8 & 87.5 [77.5, 97.5] & -- \\
Strain + clinical, logistic         & 90.0\,$\pm$\,0.0 & 90.0 [80.0, 97.5] & $+2.5$ [$0.0$, $+7.5$] \\
\bottomrule
\end{tabular}}
\end{table}

\subsection{Targeted Tests Where Motion Should Add Information}
\label{sec:targeted}
\textbf{ED-only indices.} Removing EF and LV ESV (which require the ES segmentation) lowers clinical-only 4-class accuracy from $92.9\pm0.7$ to $87.5\pm2.2\%$ (3 seeds). Late fusion with the motion-only models gives $86.2\pm1.2\%$ (graph) and $85.0\pm2.2\%$ (CNN). Although the motion branch observes the ED-to-ES deformation, late fusion does not recover the accuracy lost by removing these indices.

\textbf{DCM vs MINF} (Table~\ref{tab:dm}). Both classes require EF $<40\%$, and only MINF requires abnormal myocardial contraction. LV EDV (not indexed to body surface area) is $284.6\pm47.8$\,mL in DCM and $172.2\pm42.7$\,mL in MINF. The learned joint model changes accuracy by $-2.5$ points. Regional strain alone matches clinical-only ($87.5\%$), but adding it to the clinical indices changes accuracy by $+2.5$ points [$0.0$, $+7.5$] ($p=1.00$; one subject; not statistically significant), consistent with the small motion gain reported for this pair in \cite{zheng2019explainable}. The strain features correlate with EF (Spearman $\rho=+0.78$ for GRS, $+0.69$ for the sector s.d.\ of circumferential strain, $-0.58$ for GCS). This sector s.d.\ is higher in MINF than in DCM (one-sided, rank-biserial $r=+0.75$, $p<0.001$), but not after division by $|\mathrm{GCS}|$ ($r=+0.18$, $p=0.18$; post hoc), which is consistent with a dependence on global strain magnitude.

\begin{table}[t]
\centering
\caption{Zero-shot transfer from ACDC (3-class) to M\&Ms: balanced accuracy (\%), mean\,$\pm$\,s.d.\ over 3 seeds ($^\dagger$5 seeds). Five of the ten motion configurations are shown; ranges in the text cover all ten.}
\label{tab:mnm}
\setlength{\tabcolsep}{3pt}
\resizebox{\columnwidth}{!}{%
\begin{tabular}{lccc}
\toprule
Model & All ($n{=}141$) & Siemens (67) & Philips (74) \\
\midrule
Clinical-only, logistic, no height/weight$^\dagger$ & 63.0\,$\pm$\,0.0 & 78.9\,$\pm$\,0.0 & 46.0\,$\pm$\,0.0 \\
Clinical-only, MLP, all 8                        & 51.4\,$\pm$\,2.9 & 63.0\,$\pm$\,2.3 & 41.1\,$\pm$\,3.0 \\
Joint, no phase                                  & 61.8\,$\pm$\,0.8 & 66.9\,$\pm$\,2.2 & 51.2\,$\pm$\,0.1 \\
Joint, \texttt{dt\_scale}                        & 65.5\,$\pm$\,0.7 & 80.7\,$\pm$\,1.2 & 48.7\,$\pm$\,1.3 \\
Joint, \texttt{dt\_scale}, shuffled              & 65.3\,$\pm$\,2.6 & 76.8\,$\pm$\,2.5 & 50.8\,$\pm$\,2.7 \\
Joint, \texttt{dt\_scale}, constant              & 64.0\,$\pm$\,2.8 & 78.9\,$\pm$\,4.2 & 48.3\,$\pm$\,2.0 \\
Motion-only, no phase                            & 60.6\,$\pm$\,3.5 & 65.8\,$\pm$\,6.3 & 50.8\,$\pm$\,2.0 \\
\bottomrule
\end{tabular}}
\end{table}

\subsection{Zero-Shot Transfer to M\&Ms}
\label{sec:shift}
Overall balanced accuracy is $58.1$--$65.6\%$ for the ten motion configurations and $63.0\%$ for clinical-only (Table~\ref{tab:mnm}), carried by the Siemens subset ($61.1$--$80.7\%$; clinical-only $78.9\%$). On the Philips subset, all twelve configurations, including clinical-only, reach $41.1$--$51.5\%$. The \texttt{dt\_scale} gain over no phase ($+2.0$ points [$-3.3$, $+7.6$], $p=0.54$) recurs with shuffled ($65.3\%$) and constant ($64.0\%$) phase, so it is not attributable to physiological timing. Vendor is confounded with center and pathology, thus per-vendor values are descriptive.

\section{Discussion and Conclusion}
\label{sec:conclusion}
If ACDC labels are determined by volumetric indices, three predictions follow, and our results are consistent with all three: (1) adding motion does not significantly raise accuracy, (2) the discriminative motion features correlate with EF, and (3) regional strain does not significantly reduce the DCM vs MINF error. This agrees with clinical evidence that deformation carries information beyond EF mainly before EF declines or when it is preserved \cite{mihos2025speckle,romano2020feature}. The results do not exclude a contribution of motion (DCM vs MINF: $+2.5$ points, 95\% CI up to $+7.5$). However, accuracy on volumetrically defined labels cannot separate this contribution from that of the indices, and selecting the epoch by held-out-fold accuracy can report a motion model's accuracy from an epoch before motion is learned. Therefore, motion representations should be evaluated on outcomes not defined by these indices (such as survival \cite{bello2019deep}), against a clinical-only baseline with epoch selection that does not use the test fold. This study is limited by early stopping of the motion models on the test fold, by 2D slice-wise processing, by strain that has not been validated against tagged MRI, and by a motion representation that was not validated directly, since the graph encoder samples fixed ED contour positions and 52 of 100 ACDC sequences omit ES (the CNN encoder and the regional strain features partially address this limitation but do not resolve it).

\section{Compliance with Ethical Standards}
\label{sec:ethics}
This research study was conducted retrospectively using human subject data made available in open access by ACDC~\cite{bernard2018deep} and M\&Ms~\cite{campello2021multi}. Ethical approval was not required as confirmed by the license attached with the open access data.

\section{Acknowledgments}
\label{sec:ack}
The authors have no competing interests to declare that are relevant to the content of this article.

\bibliographystyle{IEEEtran}
\bibliography{IEEEtranctl,mybibliography,newrefs}

\end{document}